\pdfoutput=1

\documentclass[11pt]{article}

\usepackage{emnlp2021}

\usepackage{times}
\usepackage{latexsym}

\usepackage[T1]{fontenc}

\usepackage[utf8]{inputenc}

\usepackage{microtype}

\usepackage{amsmath}
\usepackage{amsfonts}
\usepackage{amsthm}
\usepackage{amssymb}
\usepackage{graphicx}
\usepackage[ruled,linesnumbered]{algorithm2e} 
\usepackage{booktabs}
\usepackage{stackengine}

\DeclareMathOperator{\argmax}{argmax}

\title{Longitudinal Citation Prediction using Temporal Graph Neural Networks}

\author{
Andreas Nugaard Holm$^1$\footnote{Contact Author}\and
Barbara Plank$^2$\and
Dustin Wright$^{1}$\and
Isabelle Augenstein$^1$\\
$^1$Department of Computer Science, University of Copenhagen\\
$^2$Department of Computer Science, IT University of Copenhagen\\
\{aholm, dw, augenstein\}@di.ku.dk,
bapl@itu.dk
}

\begin{document}
\maketitle
\begin{abstract}
Citation count prediction is the task of predicting the number of citations a paper has gained after a period of time. Prior work viewed this as a static prediction task. As papers and their citations evolve over time, considering the dynamics of the number of citations a paper will receive would seem logical. Here, we introduce the task of sequence citation prediction. The goal is to accurately predict the trajectory of the number of citations a scholarly work receives over time. We propose to view papers as a structured network of citations, allowing us to use topological information as a learning signal. Additionally, we learn how this dynamic citation network changes over time and the impact of paper meta-data such as authors, venues and abstracts. To approach the new task, we derive a dynamic citation network from Semantic Scholar spanning over $42$ years. We present a model which exploits topological and temporal information using graph convolution networks paired with sequence prediction, and compare it against multiple baselines, testing the importance of topological and temporal information and analyzing model performance. Our experiments show that leveraging both the temporal and topological information greatly increases the performance of predicting citation counts over time.
\end{abstract}

\section{Introduction}
The problem of predicting citation counts of papers has been a long-standing research problem. Predicting citation counts allows us to better understand the relationship between a paper and its impact.
However, prior research has viewed this as a static prediction problem, i.e. only predicting a single citation count at a static point in time. 
This ignores the natural development of the data as new papers are being published. Here, we propose to view the problem as a sequence prediction task, with models then having the ability to capture the evolving nature of citations.
\begin{figure}[t]
    \includegraphics[width=0.32\linewidth]{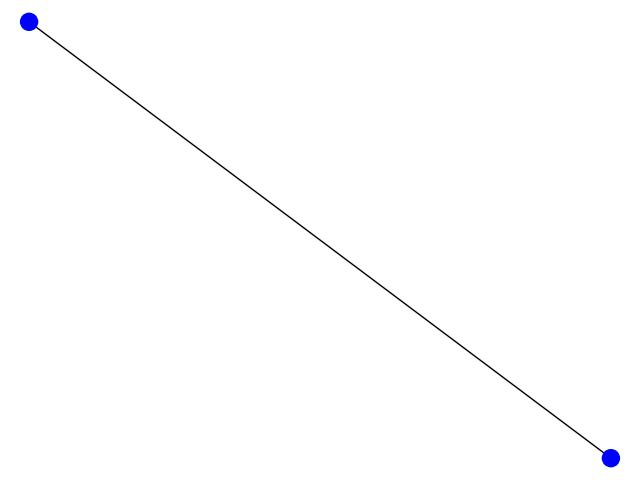}
    \includegraphics[width=0.32\linewidth]{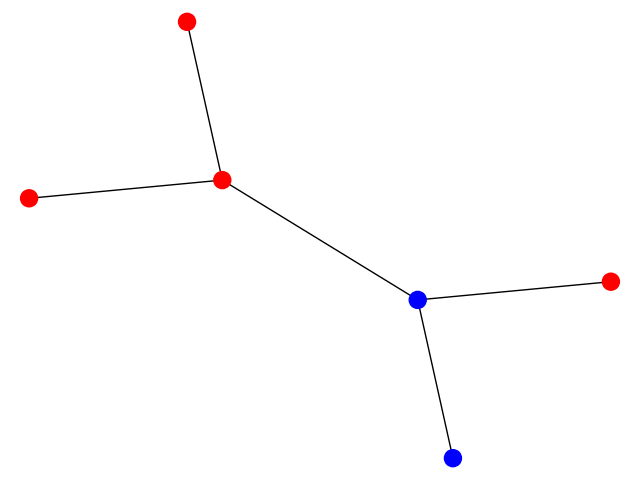}
    \includegraphics[width=0.32\linewidth]{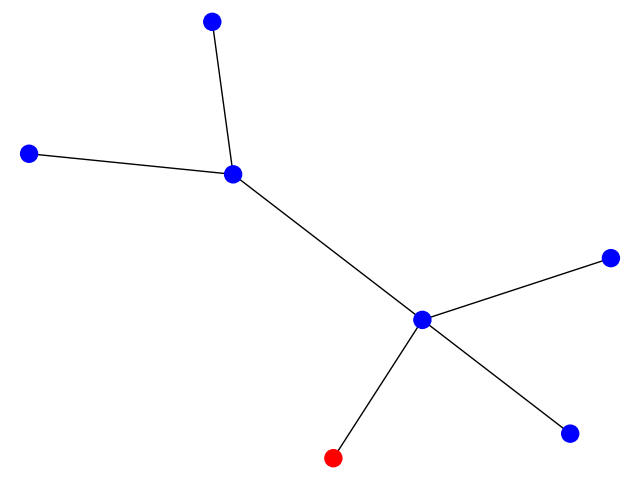}
    \caption{Illustration of the development of the dynamic graph through three time steps. Each node represents a paper; edges are citations between papers. Red nodes represent new papers in the current time step.}
\end{figure}
This, in turn, requires a dataset to contain the papers' citation counts over a period of time, which adds a temporal element to the data, which can then be encoded by sequential machine learning models, such as Long short-term memory models (LSTM)~\cite{hochreiter_long_1997}. Additionally, scholarly documents exhibit a natural graph-like structure in their citation networks. Given recent developments in modeling such data~\cite{zhou_graph_2020,wu_comprehensive_2021} and prior research showing that modeling input as graphs can be beneficial, we hypothesize that modeling a paper's citation network is useful for predicting citation counts over time. 

In this paper, we consider citation networks, a dynamic graph which evolves over time as new citations and papers are added to the network. Leveraging the structured data in the graph allows us to discover complex relationships between papers.
We want to tap into that knowledge and treat the citation data as a network, such that we can further exploit topological information and not just temporal information. By doing so, we investigate the hypothesis of paper citation counts being correlated with features such as authors, venue, and topics.

We use the well-established Semantic Scholar dataset~\cite{ammar_construction_2018} to construct our citation network. Its meta-data allows us to construct a dynamic citation network which covers a $42$ year time-line, with an updated graph for each year. The Semantic Scholar dataset's meta-data also contains information about each paper's authors, venue, and topics, allowing us to study the correlation between these features and the citation count of a paper when considering the evolving nature of the citation network. The correlation between these features and citation counts is well-known and studied by prior work~\cite{yan_citation_2011}. Prior studies show that citations are correlated and there is a strong correlation between features such as authors, but are limited by only predicting a single citation, and not predicting the natural evolution of a papers growth. 

We propose to use the constructed dynamic citation network (see Section~\ref{ssec:graph_construction}) to predict the trajectory of the number of citations papers will receive over time, a new sequence prediction task introduced in this work.
Furthermore, we propose an encoder-decoder model to solve the proposed task, which uses graph convolutional layers~\cite{kipf_semi-supervised_2017} to exploit the graphs' topological features and an LSTM to model the temporal component of the graphs. We compare our model against a vanilla graph convolutional neural network (GCN) and a vanilla LSTM, which individually incorporate either the topological information or the temporal information, but not both.

\textbf{Our contributions} are as follows: 1) A dynamic citation network based on the Semantic Scholar dataset. The dynamic citation network contains $42$ time-steps, with an updated graph at each time-step, based on yearly information. 2) We introduce the task of sequence citation count prediction. 3) A novel encoder-decoder model based on a GCN and LSTM to extract the dynamic graph's topological and temporal components. 4) A thorough study of the correlation between citation counts and temporal components.

\section{Related Work}
\subsection{Citation Count Prediction}
The task of predicting a paper's citations aims to predict the number of citations which a paper has obtained either by a given year or after $n$ years. The task itself is not new and has been researched throughout the years, and multiple different approaches have been tried and shown to be effective.  Some of these studies, have focused on feature vectors~\cite{yan_citation_2011,yu_citation_2014} and explored distinct feature vectors' performance, where they primarily rely on meta-data, e.g. venue and authors. As peer review data has become available~\cite{kang_dataset_2018}, recent research has focused on using non-meta-data information, such as peer-reviews~\cite{plank_citetracked_2019,li_neural_2019} to predict a paper's citation count. 

What is common in existing research is the target task: predicting a single citation count. This citation count can be set as one of the following years, or the citation count $n$ years in the future. To predict these citation counts, we see a variety of different neural network models with distinct architectures~\cite{li_neural_2019,wen_paper_2020}, as well as papers which focus on deeper feature vector analysis, where regression models are used \cite{davletov_high_2014,yu_citation_2014}. A side effect from prior research's focus on predicting single citation counts is that the utilized citation networks are static graphs, based on paper databases such as ArnetMiner ~\cite{tang_social_2007}, Arxiv HEP-TH~\cite{manjunatha_citation_2003} and CiteSeerX~\cite{caragea_citeseerx_2014}. These static citation networks are not suitable for our proposed task because they only contain the topological information at a single point in time.

Citation networks are not exclusively used for citation count prediction. Other citation networks such as Cora~\cite{sen_collective_2008}, CiteSeer~\cite{giles_citeseer_1998} or PubMed~\cite{sen_collective_2008}, all well known benchmark graphs, are used for node classification tasks, where the task is to predict a paper's topic. These networks are provided with minimal content. They consist of an adjacency matrix, the connections between citations, and a simple feature vector for each node of either $0/1$-valued vector or a tf-idf vector, based on the dictionary of the paper content. These existing datasets do not fit our purpose, hence we derive our own, described in Sec. \ref{ssec:graph_construction}.

\section{Temporal Graph Neural Network}
Our model is an encoder-decoder model and therefore consists of two major components. The first component is the encoder, which takes an adjacency matrix of node connections and a node feature matrix as input, where the node feature matrix can e.g.\ consist of author information (illustrated in Figure~\ref{fig:architecture}). It uses the topological information from the graphs and creates feature vectors containing both the topological node features via a GCN. It should be noted that due to the use of dynamic graphs, the encoder generates a sequence of graph embeddings, one for each graph in the sequence. The second component, the decoder, utilizes the sequence of graph embeddings created by the encoder. By using an LSTM, we extract the temporal elements and create a sequence of citation count predictions (CCP) for each node in the dynamic graph.

\subsection{Problem Definition}
While the task of CCP has been researched before, in this paper, we are interested in predicting a sequence of citation counts, which to our knowledge is so far unexplored.

Let us start by introducing our graph notation. We denote our dynamic graph as $G = \{G_0 \ldots G_{T - 1}\}$, where $G_t$ is a graph, at the given time $t$. Each graph in the dynamic graph set is defined as $G_t = (V_t, E_t)$, where $V_t$ is the set of vertices at time $t$ and $E_t$ is the set of edges at time $t$. With a given dynamic graph, we aim to predict the sequence of citations for given paper. We formalize this as $y^v = \{ y^v_1  \ldots y^v_T \}$, where $y^v_t$ is the number of citations for $v_t \in V_t$ and $y^v_t = |E^v_t|$. For our proposed task, we are given the dynamic graph $G$, and are to predict the sequence of citation counts $y$.

\subsection{Topological Feature Extraction}
One of the central hypotheses we want to examine is if complex structural dependencies in a citation network can help predict the citation count of a paper. To test this, we employ a GCN to extract topological dependencies from the graphs. We choose a GCN over other methods as they work in Euclidean space, and are thus easy to use with other neural architectures such as convolutional neural networks (CNN)~\cite{wu_comprehensive_2021}.

The GCN uses the data flow between edges in the graph to create a graph embedding. As such, we can create an embedding influenced by all of the neighboring nodes in the graph.
In this, we hypothesize that there is a relationship between the number of citations a given paper receives and that of its neighbors. The connections between the papers is described by an adjacency matrix $A$. Using our notation, we describe the GCN as follows:
\begin{align}
    H^{(l + 1)}   &= \sigma \left( \tilde{D}^{-\frac{1}{2}} \tilde{A} \tilde{D}^{-\frac{1}{2}} H^{(l)} W^{(l)} \right),
\end{align}
where $\tilde{A} = A + I$; $I$ is the identity matrix (which enables self-loops in $\tilde{A}$); $\tilde{D}_{ii} = \sum_j \tilde{A}_{ij}$, $l$ is the $l$'th layer in the model; $\sigma$ is an activation function; and $H^{(l + 1)}$ is the output of the GCN layer $H^{(l)}$. We can then simplify the above equation:
\begin{align}
    H^{(l + 1)} &= \sigma \left( \hat{A}_t H_t^{(l)} W^{(l)}  \right)
\end{align}
where $\hat{A}$ is defined as $\hat{A} = \tilde{D}^{-\frac{1}{2}} \tilde{A} \tilde{D}^{-\frac{1}{2}}$ and $t$ is the time step in the dynamic graph. It should be noted that $t$ has been left out in the first equation for simplicity. We also observe here that by adding multiple GCN layers, we allow the the graph embeddings to be affected by extended neighbours.

Since we work on a dynamic citation network, we have $T$ distinct adjacency matrices, and we have to create a graph embedding for each graph in the sequence:
\begin{align}
    Z = \{Z_0 \ldots Z_T\} &= \{ f(X, A_t) \ | \ A_t \in A \},
\end{align}
where the function $f$ is the GCN network, $Z_t \in \mathbb{R}^{m \times n}$ is a single graph embedding of dimensionality $n$ with $m$ nodes, and $Z$ is the set of graph embeddings created by the GCN. It should be noted that $X$ is shown as being independent of time, which is true for some of our node embeddings. However, some of our node embeddings are based on citations, which change through time, which makes $X$ dependent on time. We will explore the distinct node embeddings in a later section. As shown in the equation, we also keep the same model over time, and do not change the GCN even though the graph changes. We instead try to generalize the model, working on all the graphs in the dynamic graph.

\subsubsection{Temporal Feature Extraction}
With the constructed graph embeddings, containing both topological information and node information. We want to extract the temporal information, which we use the sequence of graph embeddings to do. To extract the temporal information, we utilize an LSTM, where we can formalize the input and output as $Y = l(Z)$, where the function $l$ is the LSTM and $Y \in \mathbb{R}^{m*T}$ are the CCPs.

\subsubsection{Encoder-Decoder}
In the final model, we combine the GCN and LSTM in an encoder-decoder model.
The primary challenge in combining these two models though is that they operate on vastly different inputs. The GCN operates on entire graphs and needs all the nodes to appear in the graphs, including nodes which it intends to predict. The LSTM, however, does not have this requirement and can work on batches. To solve this issue in a simple yet effective approach, we embed the entire graph prior to the LSTM steps so that in the LSTM step, we can still split the data into batches for training, validation and testing. While other approaches have been researched, like embedding the GCN into the LSTM~\cite{zhao_t-gcn_2019}, we found the simple approach to perform better.

\begin{figure}[t]
    \includegraphics[width=0.85\linewidth]{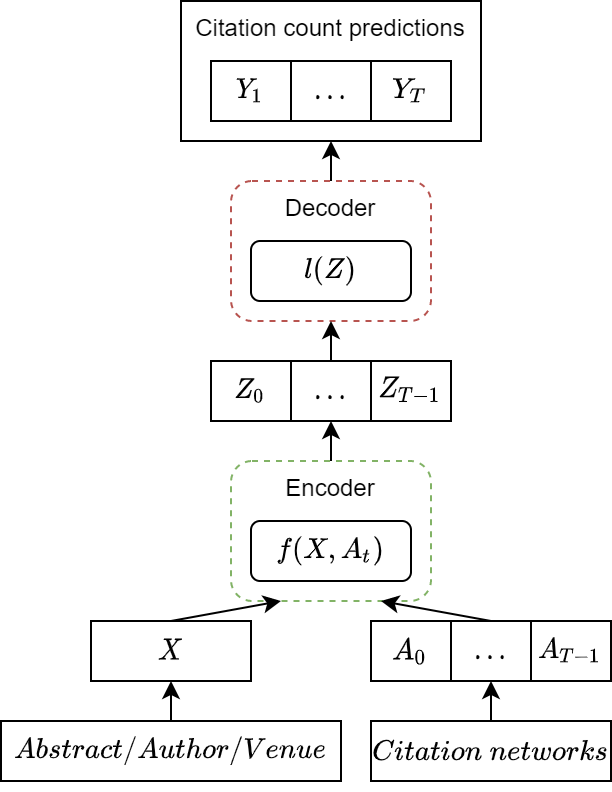}
    \centering
    \caption{Our proposed encoder-decoder model}
    \label{fig:architecture}
\end{figure}

Figure~\ref{fig:architecture} shows the architecture of our model. The GCN uses two layers to create the graph embedding. The LSTM is a single one-directional layer whose outputs are reduced to a sequence of scalars through a linear layer.

\section{Dynamic Citation Count Prediction}
\label{sec:task}
As discussed earlier, we differentiate ourselves from prior work by predicting a sequence of citation counts over time as opposed to a single final citation count. Datasets for the latter exist, but are based on paper databases. However, existing citation networks are not usable for our task due to the graph of the citation network being static in those works, i.e., the citation network does not evolve over time. Given this, we construct a dataset, where we reconstruct the citation networks, at each time-step, for the purpose of studying citation count prediction over time.

\subsection{Dataset}
The dataset which we used to create our dynamic graph is based on Semantic Scholar~\cite{ammar_construction_2018}\footnote{https://api.semanticscholar.org/}. The dataset is a collection of close to $200,000,000$ scientific papers; the size of a graph of this size requires an immense system to run experiments. To reduce the dataset to a manageable size, we only kept papers from the following venues related to AI, Machine Learning and Natural Language Processing: \textit{ACL}, \textit{COLING}, \textit{NAACL}, \textit{EMNLP}, \textit{AAAI}, \textit{NeurIPS} and \textit{CoNLL}. With the dataset only containing papers from the listed venues, we reduced the dataset's size to $47,091$ papers. Furthermore, the Semantic Scholar dataset also holds an extensive collection of meta-data for each paper. We use this meta-data to construct our dynamic graph, as well as the graph's node embeddings.

\subsection{Graph Construction}
\label{ssec:graph_construction}
With the dataset reduced to a more manageable size, we search for an ideal dynamic graph of the citation network. We define an ideal dynamic graph as the sequence of graphs which has the largest connected graph in the final graph and has the most significant increase of nodes over time. We do not use the largest connected graph at each time step, as it can trick us into selecting a sub-optimal dynamic graph. A sub-optimal dynamic graph may present itself as the largest connected graph at a point in time, but will not stay as the largest connected graph through time, and will contain less nodes through time, compared to the ideal dynamic graph. To solve the issue of being tricked into selecting a less ideal dynamic graph, we have to probe each node in the data to observe the graphs' evolution. We define probing as the process of observing the evolution of the graph connected to the probed node. This process is automatically performed on all nodes of the largest connected graph in the final step. By probing all the nodes, we can choose the sequence of graphs which contains the most nodes over time. 
In Algorithm~\ref{alg:dynamic_graph}, we describe the process in the form of pseudo-code for a more precise insight in the process of constructing the ideal dynamic graph.

\begin{algorithm}[t]
\fontsize{10}{10}\selectfont
	\KwIn{data}
	\KwOut{G}

	$connected\_graphs = \text{dict}()$ \\
	\For{$y \in \text{years}$}{
        $gs \leftarrow \text{find\_connected\_graphs}(data[y])$ \\
        $connected\_graphs[y] \leftarrow \text{sort}(gs)$ \\
	}
	
	\For{$paper \in data[\text{min}](years)$}{
	    $key\_size[paper] = 0$ \\
	    \For{$y \in \text{years}$}{
	        $best = 0$ \\
	        \For{$g \in connected\_graphs[y]$}{
	            \If{$paper \in g \ \text{and} \ |g| > best$}{
	                $best = |g|$ \\
	            }
	        }
	        $key\_size[paper] += best$ \\
	    }
	}
	$best\_paper = \argmax(key\_size)$ \\
	$G = \text{dict}()$ \\
	\For{$y \in \text{years}$}{
	    \For{$g \in connected\_graphs[y]$}{
	        \If{$best\_paper \in g$}{
	            $G[y] = g$ \\
	            $\text{break}$ \\
	        }
	    }
	}
	\caption{Dynamic Graph Construction}
	\label{alg:dynamic_graph}
\end{algorithm}

In Table \ref{table:graphs}, we show some of the properties of the last $10$ graphs in the dynamic graph. It is clear how the graph is evolving over time, as can be seen in how both the number of vertices and edges increases, and how the degree $D$ increases, indicating that the nodes in the graph obtains more citations over time. This indicates that the dynamic graph reflects the natural growth of a paper's citations.

\begin{table*}[t]
\scalebox{0.77}{
\begin{tabular}{@{}ccccccccccc@{}}
\toprule
 & 2011 & 2012 & 2013 & 2014 & 2015 & 2016 & 2017 & 2018 & 2019 & 2020 \\ \midrule
$|V|$ & $14,584$ & $16,603$ & $18,529$ & $20,760$ & $23,327$ & $26,529$ & $29,293$ & $33,759$ & $38,080$ & $38,168$ \\
$|E|$ & $103,519$ & $127,277$ & $152,869$ & $181,666$ & $217,807$ & $267,940$ & $308,186$ & $387,738$ & $475,007$ & $476,015$ \\
Mean $D$ & $7.1$ & $7.67$ & $8.25$ & $8.75$ & $9.34$ & $10.1$ & $10.52$ & $11.49$ & $12.47$ & $12.47$ \\
Max $D$ & $614$ & $761$ & $923$ & $1,072$ & $1,220$ & $1,371$ & $1,496$ & $1,763$ & $2,084$ & $2,086$ \\
Max citation count & $2,584$ & $3,110$ & $3,637$ & $4,186$ & $4,740$ & $5,403$ & $11,385$ & $20,893$ & $32,278$ & $35,200$ \\
Avg. citation count & $26.33$ & $27.48$ & $28.87$ & $30.15$ & $31.49$ & $32.94$ & $35.84$ & $38.31$ & $43.0$ & $45.41$ \\ \bottomrule
\end{tabular}
}
\centering
\caption{Key values of the graphs.}
\label{table:graphs}
\end{table*}

By only using a subset of the nodes from the full graph to construct the dynamic graph, we ablate some of the full graph's properties. One notable property of the full graph is that the citation count of a paper is tied to the degree of a node; by using a subset of the full graph this property does not hold anymore, which leads to the following definition of the size of the set of edges $y^v_t = |E^v_t|$ changing to the following for a given node $y^v_t \geq |E^v_t|$. Another important point is that removing edges from the graph removes some of the information contained in the full graph (e.g. links to papers in other fields). Such edges are usually connected to more prominent papers because it is often the high impact papers, which obtain citations from papers outside the main field.

\subsection{Feature Generation}
\label{ssec:feature_gen}
The created dynamic graph nodes are not dependent on a set of specific features, and we can therefore select and create a set of features for each node containing our desired information. With a wide variety of meta-data fields available, we created a set of distinct features which we used for our predictions. Furthermore, we studied how each of these features affect the performance of the model.

The choice of using authors and venues as features for our model is based on the hypothesis that authors listed on a paper have a major impact on the number of citations gained. We assume the same goes for venues: if a paper is published at a more highly ranked venue, it is more likely to gain a large amount of citations compared to a paper published at a lower ranking venue. We further motivate the choice of these two features based on prior work~\cite{yan_citation_2011},  who shows that author rank and venue rank are indeed two of the three features that are most predictive. We motivate the choice of using the abstract based on the assumption that the abstract of a paper contains information on the topics discussed in the paper, which can be used to identify if paper's topic is currently popular~\cite{gerrish_language-based_2010}. We further motivate the choice of using author and venue rank, as prior work shows them to be the most descriptive features~\cite{yan_citation_2011}. The following sections provide short descriptions of the meta-data used to create these feature vectors and how each of them is calculated.

\noindent\textbf{Abstract}:
To base our model on more than meta-data, we use the abstract of the papers to create a feature vector. To create an embedding of the abstract, we utilize BERT~\cite{devlin_bert_2019}, specifically the pre-trained SciBERT~\cite{beltagy_scibert_2019} model. SciBERT is a contextualized embedding model trained using a masked language modeling objective on a large amount of scholarly literature. Representations from SciBERT have been shown to be useful for learning downstream tasks with scientific text, this is why we use them here. To obtain a feature vector of a given abstract, we tokenize the abstract text and pass this through SciBERT. SciBERT prepends a special \texttt{[CLS]} token for performing classification tasks, so we use the output representation of this token as the final feature vector for an abstract.

\noindent\textbf{Author rank}:
To include the author information, we created a feature vector which ranks the authors based on their number of citations sorted by highest to lowest. Due to many authors having the same amount of citations, we allow authors to be of the same rank. As the final step for the feature calculation, we normalize the rankings by $X' = \frac{X - X_{\text{min}}}{X_{\text{max}} - X_{\text{min}}}$.

\noindent\textbf{Venue rank}:
Together with the author rank, we also hypothesize that the venue has an impact on the number of citations of a paper. Therefore, we also created a feature ranking for the venues. The feature is calculated identically to the author rank. It should be mentioned that the meta-data contains a high amount of different labels for each of the venues which we are using. We reduce all the different labels of the same venue down to a single label for each venue, but keep each venue separated by year.

\section{Experiments}
In this section we introduce our experiments, evaluate the performance of our model, and explore the importance of exploiting topological and temporal information.

\subsection{Data}
We use the constructed dynamic graph for our experiments and test each of the three distinct feature vectors. A detailed description of the feature vectors and the dynamic graph's construction can be found in Section~\ref{sec:task}. We split our data into a training, validation, and test set, with the following splits: $60\%$, $20\%$, and $20\%$. With the splits, we achieve a training set consisting of $22,900$, and a validation and test set of $7,634$. The training, validation and test sets are generated randomly, but are kept fixed throughout the experiments.

Due to the large number of time-steps in the dynamic graph, we chose to create two different setups for our experiments. One which uses the last $10$ years and another, which uses the last $20$ years of the dynamic graph. We use the later years in the dynamic graph as these years contain the most papers and the graph has evolved the most.

While not mentioned in Section~\ref{ssec:feature_gen}, we perform some further pre-processing of the data. For the feature vectors of author rank and venue rank, we perform a normalization of the values. We also perform pre-processing of the labels due to the high fluctuation of the number of citations. We take the $log(c + 1)$ of the citation of a paper as the labels \cite{maillette_de_buy_wenniger_structure-tags_2020}. Taking the log of the citation increases the stability of the model during training.

\subsection{Experimental Setup}
We perform experiments with three distinct models: 1) Our proposed model, consisting of a GCN and LSTM; 2) a standard LSTM; 3) a standard GCN. All hyper-parameters are shared across the models. 

For our selected models, we used the well-known Adam~\cite{kingma_adam_2015} optimizer, with a learning rate of $0.001$. For the GCN we used two layers, with each layer consisting of $256$ hidden units. Both the GCN and the GCN with LSTM used this setup. The LSTM was set to have a single uni-directional layer of $128$ hidden units, with the output being reduced to $1$ dimension by a linear layer. For the models using an LSTM, we its batch size to $256$. We ran the models for $1000$ epochs and if no update to the best validation score have been observed over $10$ epochs, we terminate the training early. As mentioned, we used SciBERT to encode the abstracts, with an output vector of size $768$. The models have been run using random seeds, and each of the experiments have been executed $10$ times. In the results section, we report the mean and the standard deviation of the $10$ runs.

To see if our models were learning to predict the citation counts, we created a simple deterministic model to compare against. The model is based on predicting the mean citation count of the training and validation at each time step.

\subsection{Evaluation Metric}
\begin{table*}[t]
\centering
\scalebox{0.8}{
\begin{tabular}{@{}rrrrr@{}}
\toprule
 & GCN + LSTM & LSTM & GCN \\ \midrule
Abstract & 0.8284 $\pm$ 0.0162 & 1.0164 $\pm$ 0.0140 & 1.279 $\pm$ 0.1350 \\
Author & 0.7477 $\pm$ 0.0166 & 1.0184 $\pm$ 0.0273 & 1.1089 $\pm$ 0.0357 \\
Venue & 0.9259 $\pm$ 0.1161 & 1.0414 $\pm$ 0.0197 & 1.0828 $\pm$ 0.0030 \\
Author $+$ Venue & 0.7572 $\pm$ 0.0131 & 1.0186 $\pm$ 0.0240 & 1.1248 $\pm$ 0.0271 \\
All & 0.7940 $\pm$ 0.0138 & 1.0152 $\pm$ 0.0157 & 1.3115 $\pm$ 0.1681 \\ \bottomrule
\end{tabular}
}
\caption{The performance of our 3 models over a $10$ year period. The results are reported as the MAE of the log citations. For the $10$-year period, our deterministic approach have a MAE of $1.6378$.}
\label{table:results_10}

\scalebox{0.8}{
\begin{tabular}{@{}rrrrr@{}}
\toprule
 & GCN + LSTM & LSTM & GCN \\ \midrule
Abstract & 0.8001 $\pm$ 0.0147 & 1.0149 $\pm$ 0.0414 & 1.6690 $\pm$ 0.4404 \\
Author & 0.7462 $\pm$ 0.0911 & 1.0179 $\pm$ 0.0536 & 1.3756 $\pm$ 0.0334 \\
Venue & 0.8525 $\pm$ 0.1348 & 1.0156 $\pm$ 0.0388 & 1.3212 $\pm$ 0.0039 \\
Author $+$ Venue & 0.7515 $\pm$ 0.0889 & 1.0132 $\pm$ 0.0480 & 1.3598 $\pm$ 0.0461 \\
All & 0.7803 $\pm$ 0.0167 & 1.0165 $\pm$ 0.0383 & 1.5177 $\pm$ 0.1892 \\ \bottomrule
\end{tabular}
}
\caption{The performance of our 3 models over a $20$ year period. The results are reported as the MAE of the log citations. For the $20$-year period, our deterministic approach have a MAE of $2.0796$.}
\label{table:results_20}
\end{table*}

To evaluate the performance of the models, we measure the mean absolute error, defined as
\begin{align}
    MAE &= \frac{1}{N} \sum_{i=1}^N |y - \hat{y}|,
\end{align}
where $Y$ are the citation counts and $\hat{Y}$ are the predicted values. We also use the MAE to optimize the model. We chose to use MAE, instead of mean squared error (MSE), to mitigate outlier papers which have a high amount of citations. We additionally use MAE as the training objective for the same reason.

\subsection{Results}
As previously mentioned, we ran our experiments on dynamic graphs of $10$ years and $20$ years. The results of the $10$ year experiment is shown in Table~\ref{table:results_10}, and the results of the $20$ years experiment is shown in Table~\ref{table:results_20}. In both of the experiments we see that our models outperform the simple deterministic approach.

\begin{figure*}[t]
\centering
\begin{minipage}{.5\textwidth}
  \centering
  \includegraphics[width=1.0\linewidth]{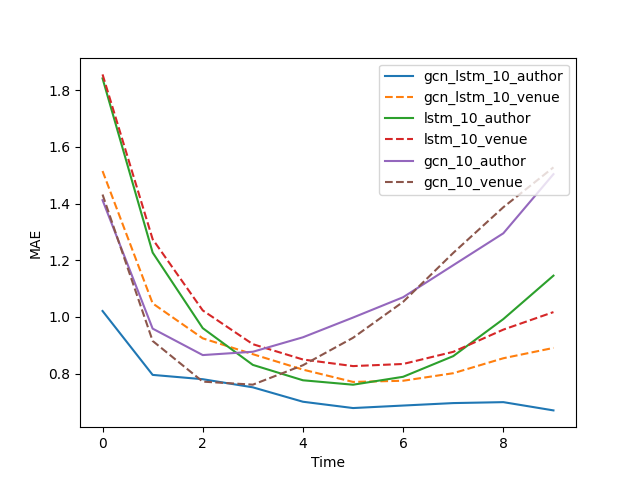}
  \includegraphics[width=1.0\linewidth]{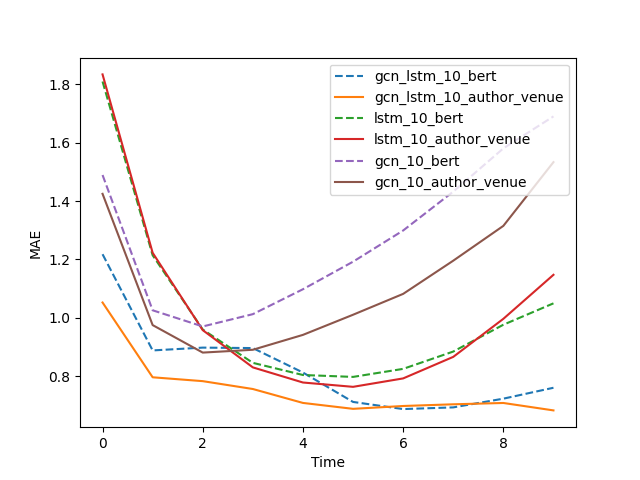}
\end{minipage}%
\begin{minipage}{.5\textwidth}
  \centering
  \includegraphics[width=1.0\linewidth]{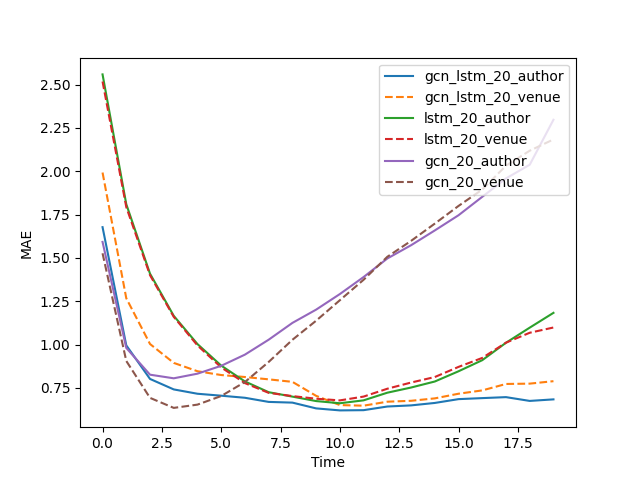}
  \includegraphics[width=1.0\linewidth]{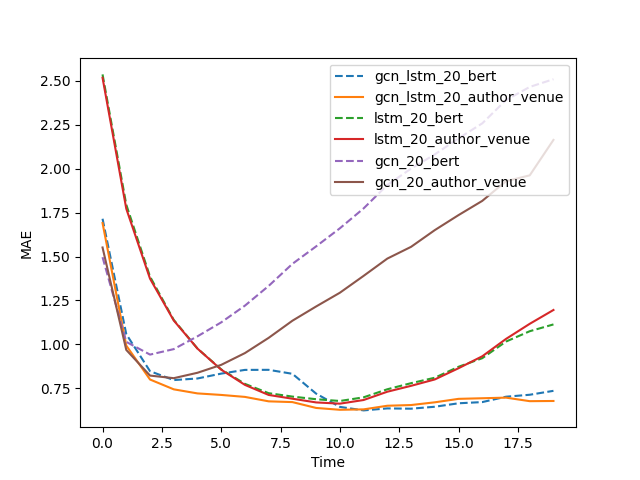}
\end{minipage}
\caption{The plots show the MAE at each time-step for our experiments, where \textbf{left} show the MAE for our $10$ year experiments and \textbf{right} show the MAE at each time-step for our $20$ year experiments, where the $x$-axis shows the time and $y$-axis the MAE.}
\label{fig:mae_time}
\end{figure*}

By inspecting the results, one can clearly observe that the GCN-LSTM has the best performance among the three models. We further observe that the GCN-LSTM improves on the performance of the pure GCN and LSTM individually, indicating that it learns from both the temporal and the topological information provided by the dynamic citation network. Furthermore, the GCN increases in error going from a $10$ year interval to a $20$ year interval, where we see the other models slightly improve. To further study this, we plot the error of the different time steps in Figure \ref{fig:mae_time}, which show the models' performances over time. By inspecting the plots, we observe a trend of the pure models i.e. the GCN and LSTM models, struggle and deteriorate over time,  compared to the combined GCN-LSTM model, which keeps improving over time until it starts plateauing. Comparing the $10$-year and $20$-year plots, one can observe that the deterioration continues, where the $10$-year plot stops. It can also be seen, that the GCN-LSTM keeps improving up until year $10$, where it levels out. All of the models decrease drastically in error up until two time-steps; afterward, the pure models start deteriorating.

\subsection{Discussion} 
Tables~\ref{table:results_10} and~\ref{table:results_20} show the impact of single feature types. We hypothesize that author information is very predictive, as shown by prior work. Inspecting the results from the different feature ablations, we can observe that the author features performs the best, confirming our hypothesis. Figure~\ref{fig:mae_time} further confirms this, showing that large parts of the gain of the model over time stems from author information.

The feature vector created by the venues performs the worst in both experiments. We hypothesize that the venues' performance could be increased if a more generalized notation for venue meta-data were available, as these are noisy (also due to OCR errors) have many spelling variants.

To further study the feature vectors, we calculate the average MAE for each distinct author and venue, where we use the predictions made by the GCN-LSTM, trained on the author feature vectors over $20$ years. We show the result of the venues in Table~\ref{table:mae_venue}. One can observe that the difference between the top and the bottom venue is drastically lower than the difference between the top and bottom author. This further indicates that the author features is a strongly predictive feature for citation counts.

\setlength{\tabcolsep}{4pt}
\begin{table}[h]
\fontsize{10}{10}\selectfont
    \centering
    \begin{tabular}{@{}lllll@{}}
    \toprule
     & Venue & MAE & Avg. degree & $n$ \\ \midrule
    1 & COLING 1973 & 0.04295 & 1 & 20 \\
    2 & AAAI 2020 & 0.06397 & 4.67 & 240 \\
    3 & NAACL 2019 & 0.0863 & 15.25 & 2160 \\
    $\vdots$ &  & & & \\
    185 & ACL 1983 & 0.7714 & 2 & 20 \\
    186 & ACL 1988 & 0.7794 & 19.6 & 100 \\
    187 & EMNLP 1998 & 0.8917 & 4.5 & 40 \\ \bottomrule
    \end{tabular}
    \caption{This table shows the top $3$ and bottom $3$ venues, sorted by the mean MAE, going from lowest to highest.}
    \label{table:mae_venue}
\end{table}

We also show the average degree and the number of papers for each of the venues in Table~\ref{table:mae_venue}. With a higher representation of papers in the collection, we expect a more reliable prediction. This is indeed the case -- we observe the top venues often have a higher number of papers in their collection. To further analyse this, we observe the average degree of the papers in the collection, however, we do not notice a higher performance where the degree is higher. This indicates that the model is better at predicting papers with higher citation counts, because the degree of a node is tightly bound to the number of citations.

\section{Conclusions}
In this paper, we propose the task of citation sequence prediction. We introduce a new dataset of scholary documents for this task based on a dynamic citation graph evolving of $42$ years, starting from a single node growing to a large graph. We further study the effect of temporal and topological information, and propose a model to benefit from both information (CGN+LSTM). Our results show that utilizing both the temporal and topological information is superior to only utilizing either the temporal or topological information. Using the proposed model, we study the effect of different features, to identify which information is most predictive of a paper's citation count over time. We find author information to be the most predictive and informative over time.

In future work, the impact of training a single GCN on the dynamic graph could be explored, since the error over time of the GCN is deteriorates fast.


\bibliography{anthology,custom}

\begin{thebibliography}{23}
\expandafter\ifx\csname natexlab\endcsname\relax\def\natexlab#1{#1}\fi

\bibitem[{Ammar et~al.(2018)Ammar, Groeneveld, Bhagavatula, Beltagy, Crawford,
  Downey, Dunkelberger, Elgohary, Feldman, Ha, Kinney, Kohlmeier, Lo, Murray,
  Ooi, Peters, Power, Skjonsberg, Wang, Wilhelm, Yuan, Zuylen, and
  Etzioni}]{ammar_construction_2018}
Waleed Ammar, Dirk Groeneveld, Chandra Bhagavatula, Iz~Beltagy, Miles Crawford,
  Doug Downey, Jason Dunkelberger, Ahmed Elgohary, Sergey Feldman, Vu~A. Ha,
  Rodney~Michael Kinney, Sebastian Kohlmeier, Kyle Lo, Tyler~C. Murray, Hsu-Han
  Ooi, Matthew~E. Peters, Joanna~L. Power, Sam Skjonsberg, Lucy~Lu Wang,
  Christopher Wilhelm, Zheng Yuan, Madeleine~van Zuylen, and Oren Etzioni.
  2018.
\newblock \href {https://doi.org/10.18653/v1/N18-3011} {Construction of the
  {Literature} {Graph} in {Semantic} {Scholar}}.
\newblock In \emph{{NAACL}-{HLT}}.

\bibitem[{Beltagy et~al.(2019)Beltagy, Lo, and Cohan}]{beltagy_scibert_2019}
Iz~Beltagy, Kyle Lo, and Arman Cohan. 2019.
\newblock \href {https://doi.org/10.18653/v1/D19-1371} {{SciBERT}: {A}
  {Pretrained} {Language} {Model} for {Scientific} {Text}}.
\newblock In \emph{Proceedings of the 2019 {Conference} on {Empirical}
  {Methods} in {Natural} {Language} {Processing} and the 9th {International}
  {Joint} {Conference} on {Natural} {Language} {Processing}
  ({EMNLP}-{IJCNLP})}, pages 3615--3620, Hong Kong, China. Association for
  Computational Linguistics.

\bibitem[{Maillette~de Buy~Wenniger et~al.(2020)Maillette~de Buy~Wenniger, van
  Dongen, Aedmaa, Kruitbosch, Valentijn, and
  Schomaker}]{maillette_de_buy_wenniger_structure-tags_2020}
Gideon Maillette~de Buy~Wenniger, Thomas van Dongen, Eleri Aedmaa, Herbert~Teun
  Kruitbosch, Edwin~A. Valentijn, and Lambert Schomaker. 2020.
\newblock \href {https://doi.org/10.18653/v1/2020.sdp-1.18} {Structure-{Tags}
  {Improve} {Text} {Classification} for {Scholarly} {Document} {Quality}
  {Prediction}}.
\newblock In \emph{Proceedings of the {First} {Workshop} on {Scholarly}
  {Document} {Processing}}, pages 158--167, Online. Association for
  Computational Linguistics.

\bibitem[{Caragea et~al.(2014)Caragea, Wu, Ciobanu, Williams,
  Fernández-Ramírez, Chen, Wu, and Giles}]{caragea_citeseerx_2014}
Cornelia Caragea, Jian Wu, Alina Ciobanu, Kyle Williams, Juan
  Fernández-Ramírez, Hung-Hsuan Chen, Zhaohui Wu, and Lee Giles. 2014.
\newblock \href {https://doi.org/10.1007/978-3-319-06028-6_26} {{CiteSeerx}:
  {A} {Scholarly} {Big} {Dataset}}.
\newblock In \emph{Advances in {Information} {Retrieval}}, Lecture {Notes} in
  {Computer} {Science}, pages 311--322, Cham. Springer International
  Publishing.

\bibitem[{Davletov et~al.(2014)Davletov, Aydin, and
  Cakmak}]{davletov_high_2014}
Feruz Davletov, Ali~Selman Aydin, and Ali Cakmak. 2014.
\newblock \href {https://doi.org/10.1145/2661829.2662066} {High {Impact}
  {Academic} {Paper} {Prediction} {Using} {Temporal} and {Topological}
  {Features}}.
\newblock In \emph{Proceedings of the 23rd {ACM} {International} {Conference}
  on {Conference} on {Information} and {Knowledge} {Management} - {CIKM} '14},
  pages 491--498, Shanghai, China. ACM Press.

\bibitem[{Devlin et~al.(2019)Devlin, Chang, Lee, and
  Toutanova}]{devlin_bert_2019}
Jacob Devlin, Ming-Wei Chang, Kenton Lee, and Kristina Toutanova. 2019.
\newblock \href {http://arxiv.org/abs/1810.04805} {{BERT}: {Pre}-training of
  {Deep} {Bidirectional} {Transformers} for {Language} {Understanding}}.
\newblock \emph{arXiv:1810.04805 [cs]}.
\newblock ArXiv: 1810.04805.

\bibitem[{Gerrish and Blei(2010)}]{gerrish_language-based_2010}
Sean~M. Gerrish and David~M. Blei. 2010.
\newblock A language-based approach to measuring scholarly impact.
\newblock In \emph{Proceedings of the 27th {International} {Conference} on
  {International} {Conference} on {Machine} {Learning}}, {ICML}'10, pages
  375--382, Madison, WI, USA. Omnipress.

\bibitem[{Giles et~al.(1998)Giles, Bollacker, and
  Lawrence}]{giles_citeseer_1998}
C.~L. Giles, K.~D. Bollacker, and S.~Lawrence. 1998.
\newblock \href
  {https://pennstate.pure.elsevier.com/en/publications/citeseer-an-automatic-citation-indexing-system}
  {{CiteSeer}: an automatic citation indexing system}.
\newblock In \emph{Proceedings of the {ACM} {International} {Conference} on
  {Digital} {Libraries}}, pages 89--98. ACM.

\bibitem[{Hochreiter and Schmidhuber(1997)}]{hochreiter_long_1997}
Sepp Hochreiter and Jürgen Schmidhuber. 1997.
\newblock \href {https://doi.org/10.1162/neco.1997.9.8.1735} {Long
  {Short}-{Term} {Memory}}.
\newblock \emph{Neural Computation}, 9(8):1735--1780.

\bibitem[{Kang et~al.(2018)Kang, Ammar, Dalvi, van Zuylen, Kohlmeier, Hovy, and
  Schwartz}]{kang_dataset_2018}
Dongyeop Kang, Waleed Ammar, Bhavana Dalvi, Madeleine van Zuylen, Sebastian
  Kohlmeier, Eduard Hovy, and Roy Schwartz. 2018.
\newblock \href {http://arxiv.org/abs/1804.09635} {A {Dataset} of {Peer}
  {Reviews} ({PeerRead}): {Collection}, {Insights} and {NLP} {Applications}}.
\newblock \emph{arXiv:1804.09635 [cs]}.
\newblock ArXiv: 1804.09635.

\bibitem[{Kingma and Ba(2015)}]{kingma_adam_2015}
Diederik~P. Kingma and Jimmy Ba. 2015.
\newblock \href {http://arxiv.org/abs/1412.6980} {Adam: {A} {Method} for
  {Stochastic} {Optimization}}.
\newblock In \emph{3rd {International} {Conference} on {Learning}
  {Representations}, {ICLR} 2015, {San} {Diego}, {CA}, {USA}, {May} 7-9, 2015,
  {Conference} {Track} {Proceedings}}.

\bibitem[{Kipf and Welling(2017)}]{kipf_semi-supervised_2017}
Thomas~N. Kipf and Max Welling. 2017.
\newblock \href {https://openreview.net/forum?id=SJU4ayYgl} {Semi-{Supervised}
  {Classification} with {Graph} {Convolutional} {Networks}}.
\newblock In \emph{5th {International} {Conference} on {Learning}
  {Representations}, {ICLR} 2017, {Toulon}, {France}, {April} 24-26, 2017,
  {Conference} {Track} {Proceedings}}. OpenReview.net.

\bibitem[{Li et~al.(2019)Li, Zhao, Yin, and Wen}]{li_neural_2019}
Siqing Li, Wayne~Xin Zhao, Eddy~Jing Yin, and Ji-Rong Wen. 2019.
\newblock \href {https://doi.org/10.18653/v1/D19-1497} {A {Neural} {Citation}
  {Count} {Prediction} {Model} based on {Peer} {Review} {Text}}.
\newblock In \emph{Proceedings of the 2019 {Conference} on {Empirical}
  {Methods} in {Natural} {Language} {Processing} and the 9th {International}
  {Joint} {Conference} on {Natural} {Language} {Processing}
  ({EMNLP}-{IJCNLP})}, pages 4913--4923, Hong Kong, China. Association for
  Computational Linguistics.

\bibitem[{Manjunatha et~al.(2003)Manjunatha, Sivaramakrishnan, Pandey, and
  Murthy}]{manjunatha_citation_2003}
J.~N. Manjunatha, K.~R. Sivaramakrishnan, Raghavendra~Kumar Pandey, and
  M~Narasimha Murthy. 2003.
\newblock \href {https://doi.org/10.1145/980972.980993} {Citation prediction
  using time series approach {KDD} {Cup} 2003 (task 1)}.
\newblock \emph{ACM SIGKDD Explorations Newsletter}, 5(2):152--153.

\bibitem[{Plank and Dalen(2019)}]{plank_citetracked_2019}
Barbara Plank and Reinard~van Dalen. 2019.
\newblock {CiteTracked}: {A} {Longitudinal} {Dataset} of {Peer} {Reviews} and
  {Citations}.
\newblock In \emph{Proceedings of {BIRNDL} {ACM} {SIGIR}, {Paris}, {France},
  {July} 25, 2019}, volume 2414, pages 116--122. CEUR-WS.org.

\bibitem[{Sen et~al.(2008)Sen, Namata, Bilgic, Getoor, Galligher, and
  Eliassi-Rad}]{sen_collective_2008}
Prithviraj Sen, Galileo Namata, Mustafa Bilgic, Lise Getoor, Brian Galligher,
  and Tina Eliassi-Rad. 2008.
\newblock \href {https://doi.org/10.1609/aimag.v29i3.2157} {Collective
  {Classification} in {Network} {Data}}.
\newblock \emph{AI Magazine}, 29(3):93--93.
\newblock Number: 3.

\bibitem[{Tang et~al.(2007)Tang, Zhang, and Yao}]{tang_social_2007}
Jie Tang, Duo Zhang, and Limin Yao. 2007.
\newblock \href {https://doi.org/10.1109/ICDM.2007.30} {Social {Network}
  {Extraction} of {Academic} {Researchers}}.
\newblock In \emph{Proceedings of the 2007 {Seventh} {IEEE} {International}
  {Conference} on {Data} {Mining}}, {ICDM} '07, pages 292--301, USA. IEEE
  Computer Society.

\bibitem[{Wen et~al.(2020)Wen, Wu, and Chai}]{wen_paper_2020}
J.~Wen, L.~Wu, and J.~Chai. 2020.
\newblock \href {https://doi.org/10.1109/ICEIEC49280.2020.9152330} {Paper
  {Citation} {Count} {Prediction} {Based} on {Recurrent} {Neural} {Network}
  with {Gated} {Recurrent} {Unit}}.
\newblock In \emph{2020 {IEEE} 10th {International} {Conference} on
  {Electronics} {Information} and {Emergency} {Communication} ({ICEIEC})},
  pages 303--306.
\newblock ISSN: 2377-844X.

\bibitem[{Wu et~al.(2021)Wu, Pan, Chen, Long, Zhang, and
  Yu}]{wu_comprehensive_2021}
Z.~Wu, S.~Pan, F.~Chen, G.~Long, C.~Zhang, and P.~S. Yu. 2021.
\newblock \href {https://doi.org/10.1109/TNNLS.2020.2978386} {A {Comprehensive}
  {Survey} on {Graph} {Neural} {Networks}}.
\newblock \emph{IEEE Transactions on Neural Networks and Learning Systems},
  32(1):4--24.

\bibitem[{Yan et~al.(2011)Yan, Tang, Liu, Shan, and Li}]{yan_citation_2011}
Rui Yan, Jie Tang, Xiaobing Liu, Dongdong Shan, and Xiaoming Li. 2011.
\newblock \href {https://doi.org/10.1145/2063576.2063757} {Citation count
  prediction: learning to estimate future citations for literature}.
\newblock In \emph{Proceedings of the 20th {ACM} international conference on
  {Information} and knowledge management - {CIKM} '11}, page 1247, Glasgow,
  Scotland, UK. ACM Press.

\bibitem[{Yu et~al.(2014)Yu, Yu, Li, and Wang}]{yu_citation_2014}
Tian Yu, Guang Yu, Peng-Yu Li, and Liang Wang. 2014.
\newblock \href {https://doi.org/10.1007/s11192-014-1279-6} {Citation impact
  prediction for scientific papers using stepwise regression analysis}.
\newblock \emph{Scientometrics}, 101(2):1233--1252.

\bibitem[{Zhao et~al.(2019)Zhao, Song, Zhang, Liu, Wang, Lin, Deng, and
  Li}]{zhao_t-gcn_2019}
Ling Zhao, Yujiao Song, Chao Zhang, Yu~Liu, Pu~Wang, Tao Lin, Min Deng, and
  Haifeng Li. 2019.
\newblock \href {https://doi.org/10.1109/TITS.2019.2935152} {T-{GCN}: {A}
  {Temporal} {Graph} {Convolutional} {Network} for {Traffic} {Prediction}}.
\newblock \emph{IEEE Transactions on Intelligent Transportation Systems}, pages
  1--11.

\bibitem[{Zhou et~al.(2020)Zhou, Cui, Hu, Zhang, Yang, Liu, Wang, Li, and
  Sun}]{zhou_graph_2020}
Jie Zhou, Ganqu Cui, Shengding Hu, Zhengyan Zhang, Cheng Yang, Zhiyuan Liu,
  Lifeng Wang, Changcheng Li, and Maosong Sun. 2020.
\newblock \href {https://doi.org/10.1016/j.aiopen.2021.01.001} {Graph neural
  networks: {A} review of methods and applications}.
\newblock \emph{AI Open}, 1:57--81.

\end{thebibliography}
\bibliographystyle{acl_natbib}

\appendix

\end{document}